\documentclass[letterpaper, 11 pt, conference]{ieeeconf}

\IEEEoverridecommandlockouts                              

\usepackage{lipsum}
\usepackage{graphicx}                       
\usepackage{stfloats}
\usepackage[tight,footnotesize]{subfigure}  

\usepackage{amssymb,amsmath}
\usepackage{textcomp}
\usepackage{mdwmath}
\usepackage{mdwtab}
\usepackage{eqparbox}
\usepackage{rotating}                       
\usepackage{array}                          
\usepackage{listings}                       
\usepackage[ruled,vlined,noresetcount]{algorithm2e}
\usepackage{listings}
\usepackage[noend]{algpseudocode}

\usepackage[colorlinks,
            linkcolor=blue,
            anchorcolor=blue,
            citecolor=blue,
            bookmarks=false
            ]{hyperref}
\usepackage{cite}

\usepackage{graphicx}
\usepackage{float}
\usepackage{subfigure}
\usepackage{amsmath}
\usepackage{algpseudocode}
\usepackage{stfloats}
\usepackage{multirow} 
\usepackage{xcolor}
\usepackage{amssymb}
\usepackage{booktabs}

\title{\LARGE \bf An Enhanced LiDAR-Inertial SLAM System for Robotics Localization and Mapping}
\author{Kangcheng Liu$^{*}$, Xunkuai Zhou,  and Ben M. Chen
\thanks{$^{*}$Kangcheng Liu is the corresponding author.}}

\begin{document}

\maketitle
\thispagestyle{empty}
\pagestyle{empty}

\begin{abstract}
The LiDAR and inertial sensors based localization and mapping are of great significance for Unmanned Ground Vehicle related applications. In this work, we have developed an improved LiDAR-inertial localization and mapping system for unmanned ground vehicles, which is appropriate for versatile search and rescue applications. Compared with existing LiDAR-based localization and mapping systems such as LOAM, we have two major contributions: the first is the improvement of the robustness of particle swarm filter-based LiDAR SLAM, while the second is the loop closure methods developed for global optimization to improve the localization accuracy of the whole system. We demonstrate by experiments that the accuracy and robustness of the LiDAR SLAM system are both improved. Finally, we have done systematic experimental tests at the Hong Kong science park as well as other indoor or outdoor real complicated testing circumstances, which demonstrates the effectiveness and efficiency of our approach.  It is demonstrated that our system has high accuracy, robustness, as well as efficiency. Our system is of great importance to the localization and mapping of the unmanned ground vehicle in an unknown environment.


\end{abstract}

\section{Introduction and Related Work}


Light Detection And Ranging (LiDAR) sensor is a typical sensor that can obtain the distance of all surrounding objects with a local range \cite{liu2022datasets2}. The LiDAR-based simultaneous localization and mapping (SLAM) is of great significance to various applications, such as autonomous driving, 3D robotics grasping, search and rescue robots, and robotics surveillance as well as inspection. In the past few years, besides various of advance control techniques \cite{liu2017avoiding}, deep learning-based methods have drawn great attention because they have recreated the visual world. 
\begin{figure}[t]
\centering
\includegraphics[scale=0.25]{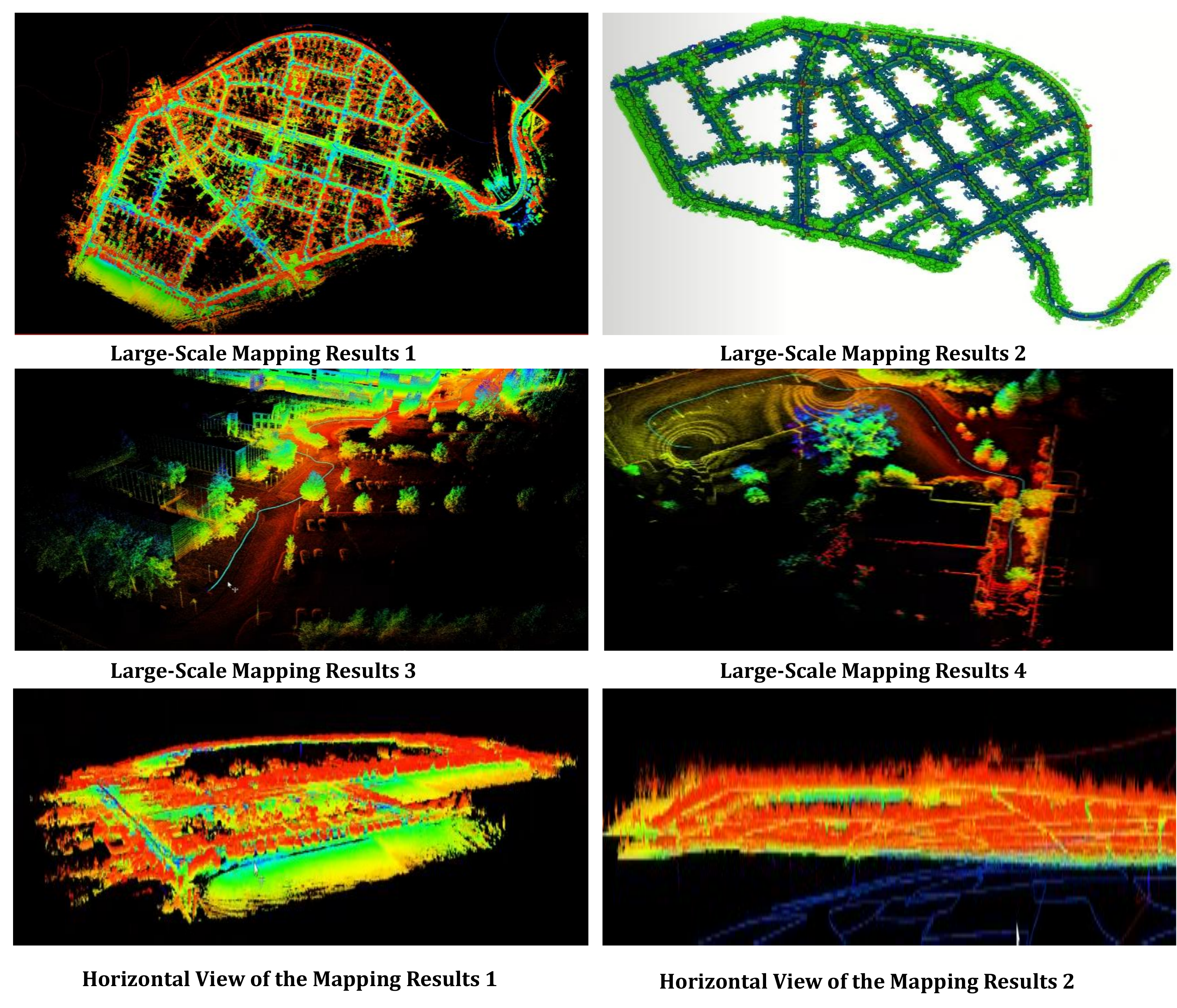}
\caption{Large-scale Mapping Results in Complex Environments}
\label{fig_overall}
\vspace{-3mm}
\end{figure}
Recently, various of LiDAR-SLAM systems have been proposed thanks to the development of large-scale LiDAR-based localization benchmarks such as KITTI  \cite{behley2019semantickitti, liu2022integratedicca, liu2022robustcyber, geiger2012we, liu2022semicyber} for autonomous driving. The gmapping-based method has been proposed \cite{grisetti2007improved}, which is a typical LiDAR SLAM based on particle filter.  The Cartographer \cite{hess2016real, liu2022robustmm} proposed by Google utilizes the Ceres solver to solve the nonlinear least square to achieve the matching in scanning. By constructing the submap and the submap-to-scan matching, the loop closure detection and the global optimization are achieved. This kind of method includes the Fast-LIO\cite{liu2022enhanced}, which utilizes the incremental KD-Tree in the front end of the odometry. The iterative Extended Kalman Filter is utilized in the backend. It has the advantages of short duration and fast calculation. The LOAM-based LiDAR-SLAM utilizes the edge and planar features of the point clouds as the front-end features. These features are stored in the map, and the point-to-line and point-to-plane distance information is utilized when matching. The high-frequency visual information and low-frequency LiDAR odometry information are both utilized to achieve the robot motion estimation \cite{zhang2015visual}. The LEGO-LOAM was proposed in \cite{shan2018lego} for fine-grained feature extraction. And real-time localization and mapping performance can be achieved by this system with acceptable accuracy in a simple environment. Recently, some methods have also been proposed to tackle the problem of limited field of views for solid-state LiDAR, such as Loam-Livox. The deep learning-based method has been proposed for feature extraction recently. However, the major problem is that non of the previous methods have included the loop closure in their system. Therefore, the LiDAR SLAM system suffers from drift and low accuracy. Also, all the above-mentioned methods utilize the Kalman Filter to calculate the odometry in the front end of the LiDAR SLAM system. However, the particle filter-based methods \cite{grisetti2007improved, liu2019deep} have not been fully explored to calculate the odometry. All in all, currently the various of research on LiDAR-based SLAM can be roughly summarized into two aspects: The first part is focused on improving the accuracy of the system. The second part is focused on improving the robustness of the system. Various of methods have been proposed to improve the localization accuracy and robustness in diverse kinds of indoor or outdoor scenarios. However, how to strike a great balance between those two factors requires more exploration.
\begin{figure*}[t]
\centering
\includegraphics[scale=0.52]{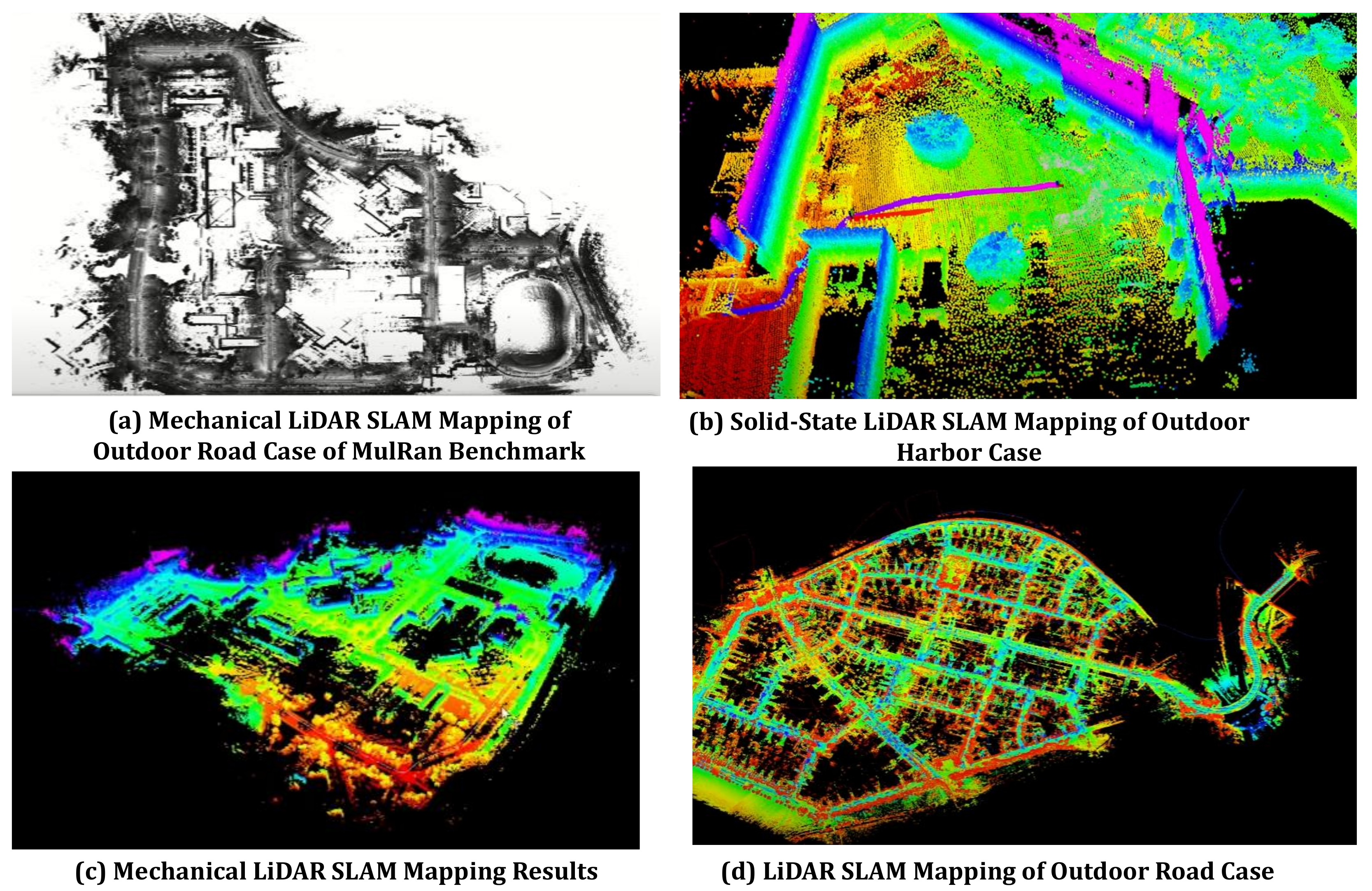}
\caption{The final demos of the Proposed LiDAR SLAM mapping results. It can be seen that on Mulran \cite{gskim-2020-mulran} benchmark, our proposed learning-based loop closure detection method can achieve satisfactory accuracy and correct large error (subfigure (a) which is the global map before proposed loop closure detection method) to a global consistent map (subfigure (b) which is the map obtained after we do the loop closure.) Subfigure (c) and Subfigure (d) demonstrate our testing results on the outdoor road case and outdoor harbor case respectively. It can be seen that global consistent mapping can be achieved both in indoor and outdoor circumstances with mechanical and solid-state LiDAR \cite{liu2022integratedar, liu2022lightar, liu2022deijcv}. The robustness of our proposed approach is demonstrated.}
\label{fig_exp}
\vspace{-1mm}
\end{figure*}
\begin{figure}[t]
\centering
\includegraphics[scale=0.35]{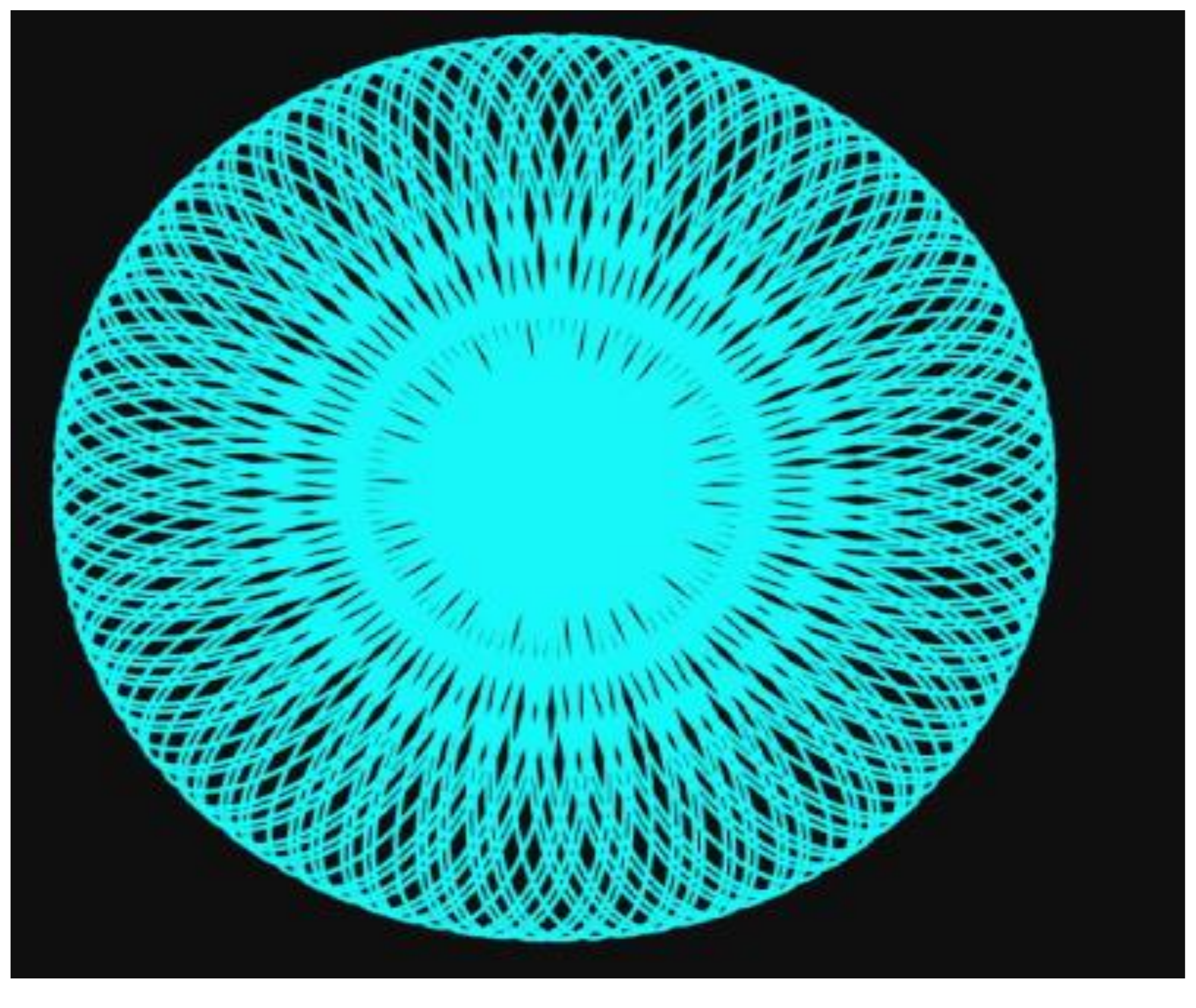}
\caption{The scanning patterns of the Solid-State LiDARs}
\label{fig_overall}
\vspace{-3mm}
\end{figure}

\begin{figure}[t]
\centering
\includegraphics[scale=0.261]{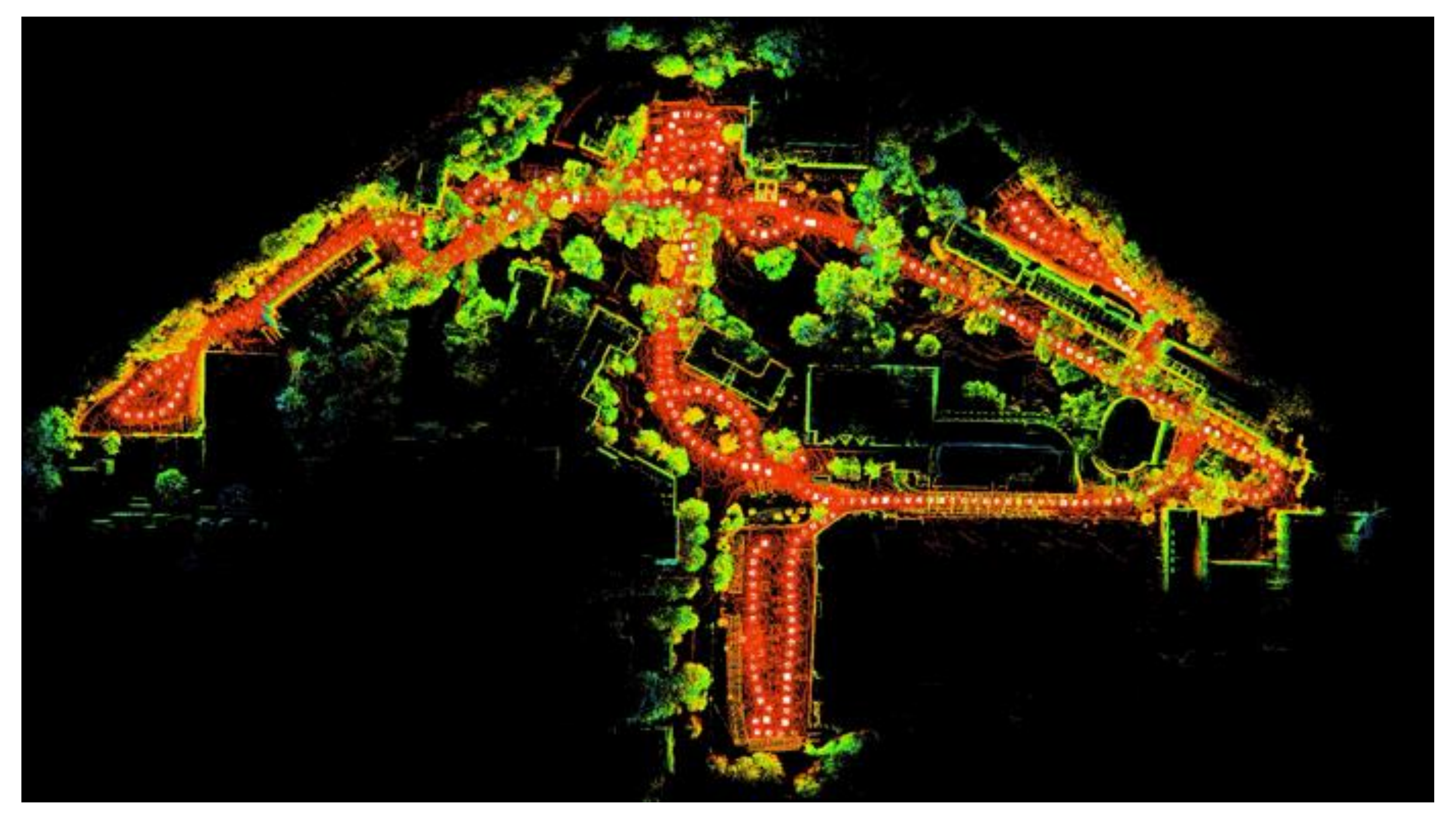}
\caption{The Mechanical LiDAR Mapping Results of the Campus Scenarios}
\label{fig_overall}
\vspace{-3mm}
\end{figure}

To tackle the challenges mentioned above, in this work, we present a particle filter-based method to solve the LiDAR inertial localization and mapping problem in unknown environments. Firstly, we have proposed an adapted particle swarm filter for both indoor and outdoor localization and mapping. We have proposed a smaller time steps in the optimization of the adapted particle swarm filter to increase the success rate and robustness in the localization. Also, we have also proposed using the resampling strategy to make the correct sampling distribution. Finally, we utilize the Extended Kalman Filter-based particle filter to conduct the state estimation. Also, we have proposed a lightweight learning-based method to find the loop closure in LiDAR SLAM. We utilize a deep neural network with a classification head to judge if two local LiDAR scans overlap. Also, we have designed a regression head for the deep neural network to predict the relative angle between the two LiDAR scans. In summary, we have the following prominent contributions:

\begin{enumerate}
\item We have proposed a particle swarm filter-based adapted framework for the localization problem. The adapted method utilizes the resampling strategy and the appropriate suggested density for optimization based on the constructed local map.

\item We have designed an effective loop closure method to improve localization accuracy and robustness. Various experiments have demonstrated the effectiveness of our proposed loop closure algorithms.

\item For the efficiency issues, we have also utilized various of nearest neighbor search algorithms for a fast calculation of the neighbors. Based on the original implementations, we have improved the speed by 5.8 times in the nearest neighbor search.

\item We have integrated the proposed method into our system both for the solid-state LiDAR of Livox-Avia as well as the mechanical LiDAR such as VLP-16. It can be demonstrated that our method has clear merits in effectiveness and efficiency. It achieves satisfactory performance for various of tested circumstances including the Hong Kong science park. Also, our method can be integrated seamlessly with the robotics safe corridor based motion planning. The robustness and accuracy of our method are demonstrated on both public datasets, as well as the real complicated indoor and outdoor circumstances.
\end{enumerate}
In the following of our work, Section II introduces our overall system framework and two of our key contributions. Next, we propose our two major contributions, the adapted particle swarm filter, and the learning-based loop closure approach. We have included our experimental results in each Subsection due to page limits.  Finally, we have briefly introduced our integrated system for local building inspections in Hong Kong.

\begin{figure}[t]
\centering
\includegraphics[scale=0.23]{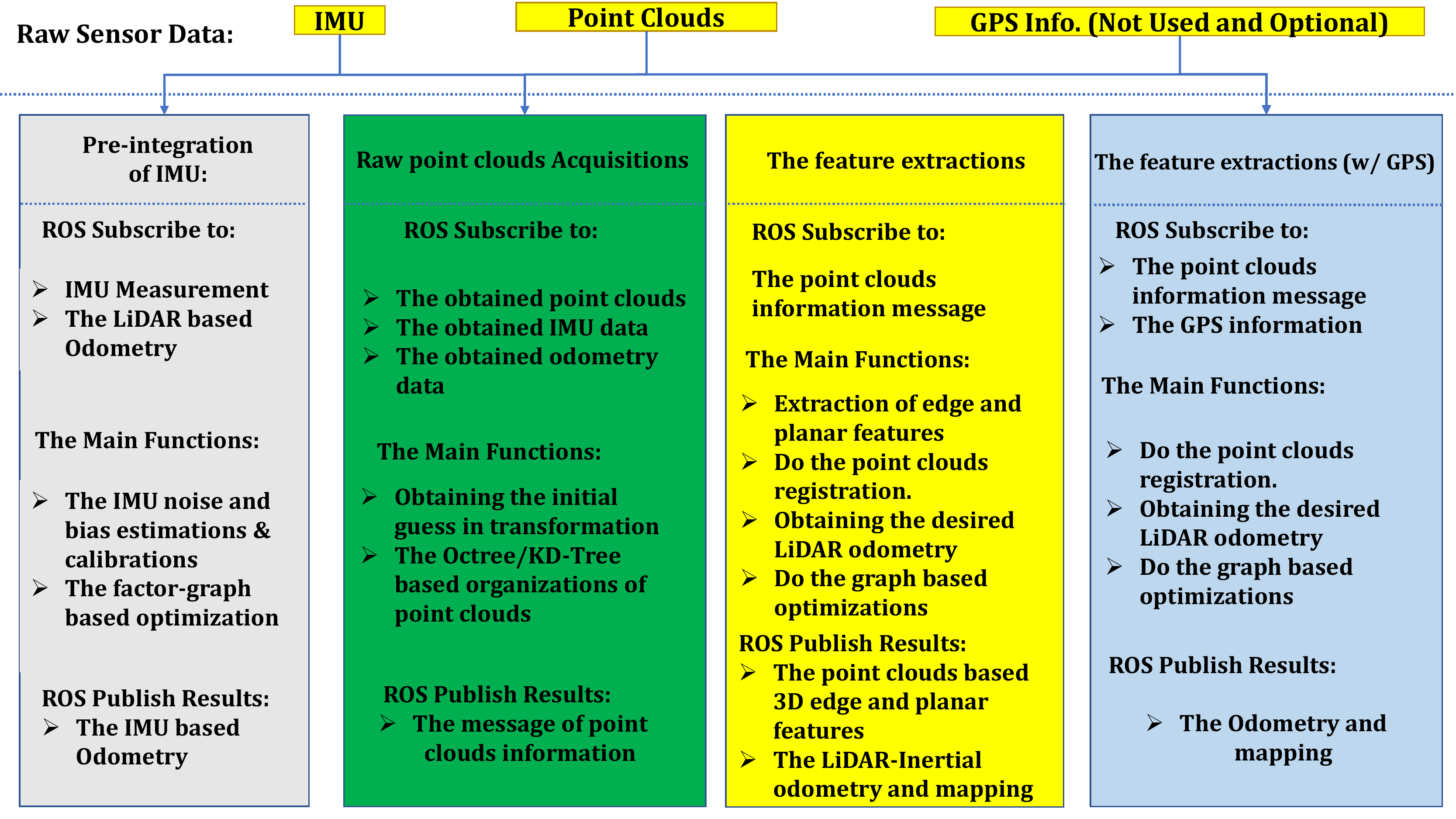}
\caption{The Final Software Framework for the proposed LiDAR-Inertial SLAM System}
\label{fig_code}
\vspace{-2mm}
\end{figure}

\begin{figure*}[t]
\centering
\includegraphics[scale=0.31]{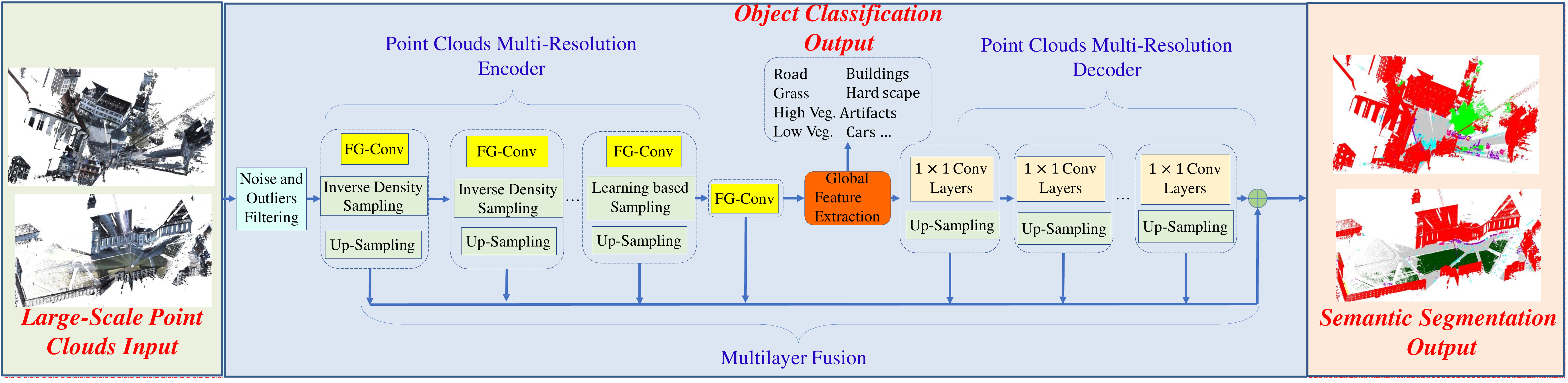}
\caption{The Overall Framework of the FG-Conv Based FG-Net Network Architecture.}
\label{fig_code}
\vspace{-2mm}
\end{figure*}
\section{Proposed Methodology}
\subsection{The Overall System Framework}
\subsubsection{The Framework of Our Proposed LiDAR-Inertial Localization and Mapping System}
The proposed localization and mapping framework is mainly based on the typical LiDAR SLAM or LiDAR Inertial system such as the LEGO-LOAM\cite{shan2018lego} and LIO-SAM \cite{shan2020lio}. Compared with their method, we have two major difference: The first is that we propose the first particle filter based which achieved better efficiency compared with previous methods such as LEGO-LOAM\cite{shan2018lego} and LIO-SAM \cite{shan2020lio}. The second is that we have proposed an effective loop closure detection network termed Loop-Net. It is demonstrated that our proposed approach can perform loop closure very effectively and the overall accuracy of the whole LiDAR SLAM system is greatly improved.  
\subsubsection{The Final Software Framework for the Proposed LiDAR-Inertial SLAM System}
The coding framework of our proposed LiDAR-Inertial SLAM is shown in the Fig. \ref{fig_code}. The running procedure of our software is summarized as follows:

1. Laser motion distortion calibration. The pre-integration is calculated by using the IMU data and IMU odometer data between the start and end of the current frame. The pose of the laser point at each moment is obtained, so as to transform it into the coordinate system of the laser at the initial moment to realize the calibration.

2. Extract features. For the current frame laser point cloud after motion distortion correction, the curvature of each point is calculated, and then the corner and plane point features are extracted.

3. Scan-to-map matching. Extract the feature points of the local key frame map, perform scan-to-map matching with the feature points of the current frame, and update the pose of the current frame.

4. Factor graph optimization or extended Kalman Filter based optimization. Add laser odometry factor, GPS factor, closed-loop factor to the graph and perform factor graph optimization, and update all keyframe poses.

5. Loop closure detection. Find candidate loop closure matching frames in historical key frames, perform scan-to-map matching to obtain the pose transformation. Then we can build the loop closure factors, and add them to the factor graph for optimization.
It should be mentioned that in order to keep the minimum requirements in our sensor suit, we have included the GPS fusion module in our software but we have not utilized the GPS information in all our experiments.

\subsection{The adapted particle swarm filter}
 \subsubsection{The principle of the particle swarm filter}
 The basic ideas of particle filter based SLAM is to use $m$ random samples $S(k)=\{s^i(k)|(X^i_r(k), w^i(k)), i=1,2,...,m\}$ It represents the confidence value of the robot at the k time step $bel(X_r(k))=p(X_r(k)), | d_{0,1,2,..,k})$. The $X_r(k)$ represents the pose value at the k step of the robot. $w_{k}$ represents the weight that is corresponding to $X_r(k)$. The $d_{0, 1, 2,.., k}$ represents the data from the time step $Z^k=\{Z(0), Z(1), .., Z(k)$, the estimated state is $X_r(k)$. The $X_r(k)$ is the pose of the mobile robot. The $X_r(k)$ is the pose of the mobile robot. The $X_r(k)$ stands for the pose of the mobile robot. $X_r(k)=(x_r(k), y_r(k),\theta_r(k))$. The system description is as follows:
 \begin{equation}
     X_r(k)=f(X_r(k-1), u(k-1), w(k-1))
 \end{equation}
 In which $u(k)$ is the motion input, such as the data from the odometry. the $w(k)$ is the process noise.
 The measurement function can be formulated as follows:
  \begin{equation}
     Z(k)=h(X_r(k-1), X_n(k), v(k))
 \end{equation}
 The $X_n(k)$ represent the route Information, while the $v(k)$ stands for the measurement noise.
 
 Then we can obtain the credibility of the robot. 
 
 \begin{equation}
     bel(X_r(k))=p(X_r(k)|Z(k), u^{k-1}, Z^{k-1})
 \end{equation}
According to the Bayes theorem, we can obtain the recursive computation, which can be formulated as:
 \begin{equation}
 \tiny
     bel(X_r(k))=\frac{p(X_r(k)|Z(k), u^{k-1}, Z^{k-1})p(X_r(k)| u^{k-1}, Z^{k-1})}{p(Z(k)| u^{k-1}, d_{0, 1, 2, 3, ..., k-1}}
 \end{equation}
According to the Markov hypothesis, the measurement $Z(k)$ is conditionally independent with the previous measurement. Denote all the previous measurements as $Z^{k}$, $Z^{k-1}=\{Z(0), ..., Z(K-1)\}$, which is merely dependant on the previous measurements. And it is only related to the pose $X_n(k)$. 
 \begin{equation}
     p(Z(k)|X_r(k))=p(X_r(k)|Z(k), u^{k-1}, Z^{k-1})
 \end{equation}
Given $X_n(k-1)$ and $u(k-1)$, the state $X_n(k)$ is independent with the previous state $X_n(k-1)$ and control $u(k-1)$, which is:
\begin{equation}
\tiny
\begin{aligned}
    bel(X_r(k))=&\frac{p(Z(k)|X_r(k), u^{k-1}, Z^{k-1}) p(X_r(k)|u^{k-1}, Z^{k-1})}{p(Z(k)|u(k-1),d_{0,1,2,...,k-1})} 
    \\
    =&\eta \times p (Z(k)|X_r(k))\times p (X_r(k)| (k-1), u(k-1)) \\
    &bel(X_r(k-1)) dX_r(k-1) 
\end{aligned}
\end{equation}
  


\begin{figure*}[ht]
\centering
\includegraphics[scale=0.52]{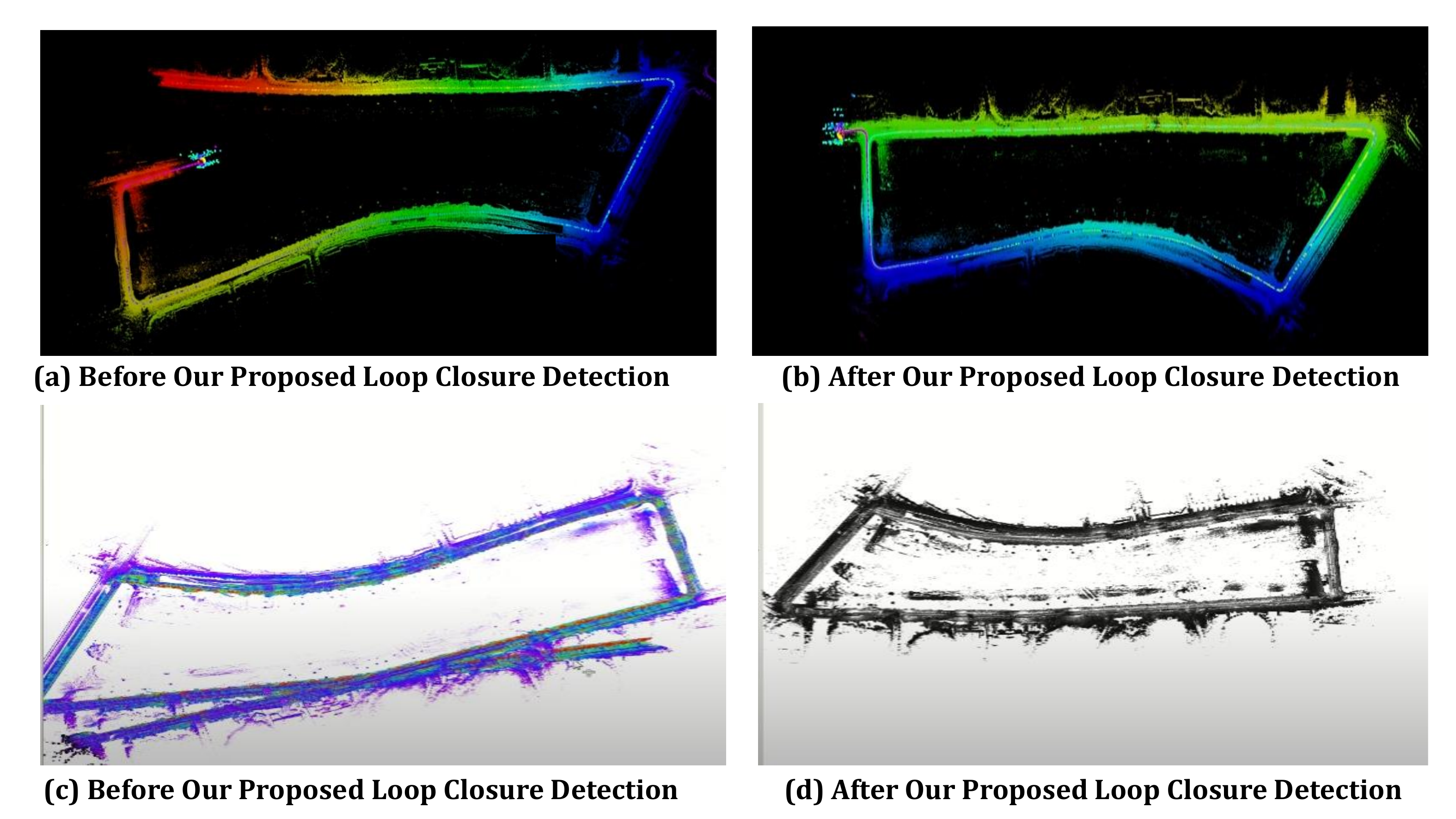}
\caption{The final demos of the Proposed LiDAR SLAM mapping results. It can be seen that on Mulran \cite{gskim-2020-mulran} benchmark, our proposed learning based loop closure detection method can achieve satisfactory accuracy and correct large error (subfigure (a) which is the global map before proposed loop closure detection method) to a global consistent map (subfigure (b) which is the map obtained after we do the loop closure.) The Subfigure (c) and Subfigure (d) demonstrate our testing results on the outdoor road case and outdoor harbor case respectively. It can be seen that the global consistently mapping can be achieved both in the indoor and outdoor circumstances with mechanical and solid-state LiDAR. The robustness of our proposed approach is demonstrated.}
\label{fig_exp}
\vspace{-3mm}
\end{figure*}









This equation is the foundation of particle swarm filter. In which $\eta = \frac{1}{p(Z(k)| u(k-1), d_{0,1,...,k-1})}$, $p(Z(k)|X_r(k)$ is the sensor model correspond to the robot. We can define $q=p(X_r(k)|X_r(k-1), u(k-1))bel(X_r(k-1))$ as the importance function. The determination of $q$ has great influence on the performance of particle filter based SLAM.

\subsubsection{Drawback of previous particle swarm filter}
We have proposed an adapted particle swarm filter for localization and mapping. Traditionally, the typical particle swarm filter \cite{grisetti2007improved} in SLAM seeks to reach a global convergence. In the ideal situations, the standard particle swarm filter based SLAM system will reach the global convergence after infinite times of iterative optimizations. And finally, we can obtain the deterministic pose of the robotic system and the deterministic mapping of the environment. However, in reality,  the mapping constructed from uncertain robotics pose is not very accurate. Simultaneously, the estimated robot pose from the uncertain environmental map is not accurate. With the movement of the robot, the accuracy in the robot pose will have great influence on the mapping accuracy and vice versa. It will result in the low accuracy and low efficiency in the robotics mapping and localization. Sometimes, some weak constrains may not be satisfied, and the final localization will not converge. In the meanwhile, the standard particle filter based SLAM takes uniformly distributed particles in the whole exploration environment. And the convergence speed will also decrease. Moreover, the particle number will also increase, which will increase the computational demand and influence the real-time performance of the whole LiDAR-Inertial SLAM system \cite{liu2022weaklabel3d, liu2022weakly}. 
\subsubsection{Smaller time step in optimization for adapted particle swarm filter}
We have proposed an adapted particle swarm filter. The adapted particle swarm filter aims at the fact that the uncertain robotics poses and uncertain mapping results. Our proposed method seeks the optimal pose for the robot at the local range \cite{liu2022enhanced, yuzhi2020legacy, yuzhi2020legacy2}. Then, the optimal pose of the robot at the local range is utilized to construct the environmental map at the local range. Our final proposed specific method can be summarized as follows: 
\begin{enumerate}
    \item Figure out the little square which is centered at the calculated robotics pose. We set the side length of the rectangle as 1cm for indoor case and 10cm for the outdoor case. 
     \item 
    We let the certain amount of particles uniformly distributed in the rectangle. After the robot moving for ten time steps, the localization is finished. 
    \item
    Then we use the LiDAR to re-scan the environment and perform the map update.    
\end{enumerate}
  In this way, utilizing the simple trick of smaller time step, and performing the optimization in a local range, we can increase the success rate and robustness in localization in the final LiDAR SLAM system. In summary, we perform particle swarm based localization based on constructed local map in a short period. Also, we let the particles within a small range with respect to the robot, which can significantly cut down the particles required. 
  \subsubsection{Proposed Approach to tackle the Degeneration Problem}

  The problem exists in the basic particle filter is the degeneration problem. The variance in weights of the particles will increase the time iteration and the degeneration is unavoidable. Through iterations, the weight of other particles will reach a small value that is negligible. Then degeneration means that large amount of time and computational resources will be wasted on the particles which have little weight. It will not only result in the waste of resources but also will influence the final estimation results. In our work, we have proposed methods to tackle the weight degeneration problems:
   \begin{enumerate}
  \item Utilizing the resampling strategy. The essence of the resampling strategy is to increase the diversity of the particles. By introducing the resampling, the diversity will become large. In our work, we utilize the multinomial resampling \cite{gordon1993novel} proposed by Gordon et al., which solves the degeneration problem in the particle filter. 
  \item We have also utilized the appropriate suggested density distribution similar to \cite{qian2015remaining}. The assumption of basic particle filter is that: The importance based resampling can sample a group of points from an appropriate posterior suggested density distribution. The suggested density distribution function guides the resampling to make the correct sample distribution. Therefore, if we can find an optimal suggested density distribution function, the quality of the final filtering can be guaranteed.
  \end{enumerate}
  
 \begin{figure*}[ht]
\centering
\includegraphics[scale=0.52]{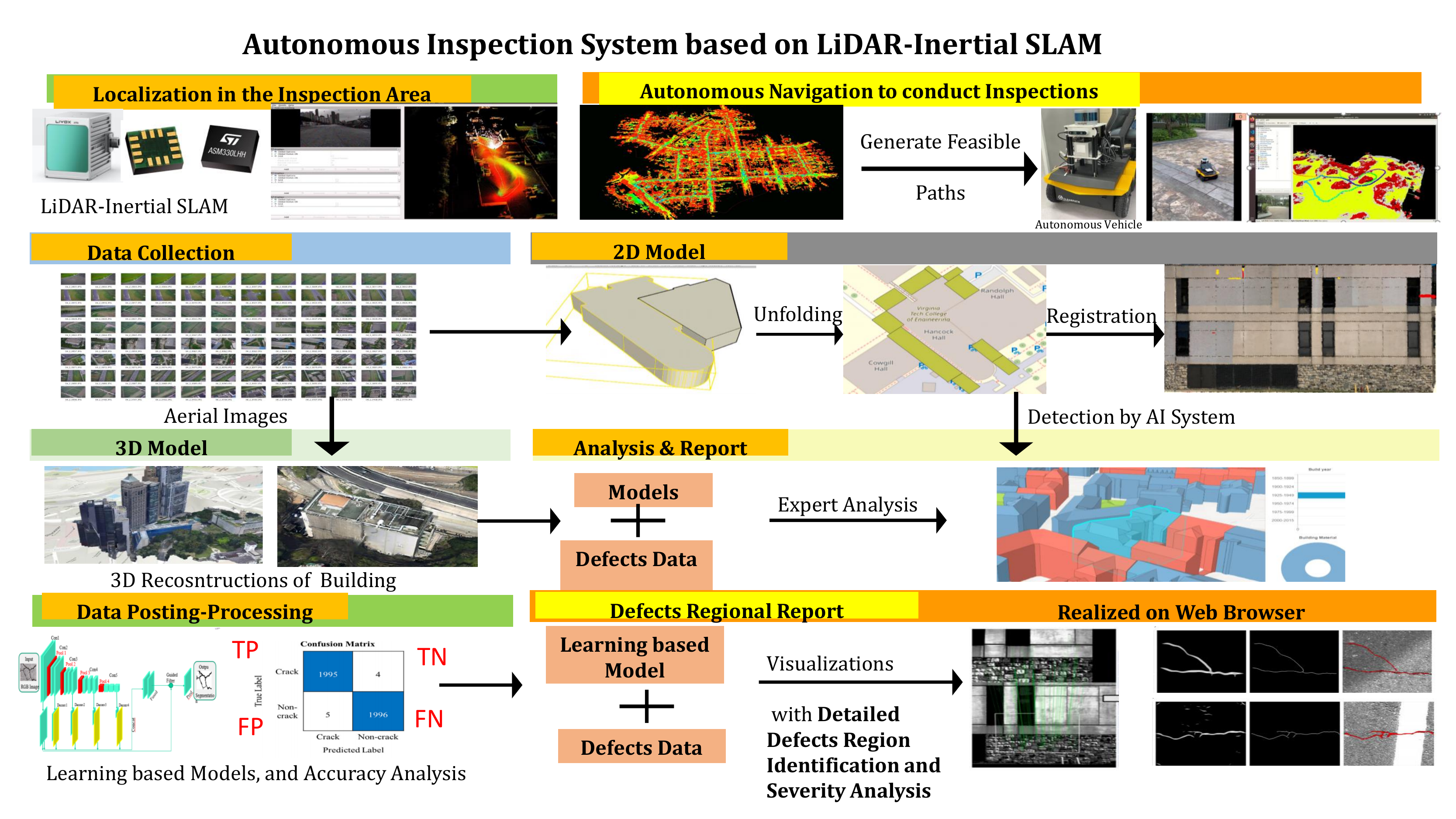}
\caption{Detailed illustration of the deployed system for UGV Inspections based on our proposed LiDAR SLAM system}
\label{fig_System_Inspection}
\vspace{-0.1cm}
\end{figure*}      
    
\section{The Learning-based Loop Closure}
Also, we have proposed a loop closure algorithm for the SLAM system based on deep learning. By our proposed loop closure detection techniques, the features of similar places can be effectively compared, and the loop closure can be effectively detected. Also, our proposed network architecture is very lightweight and appropriate for real-time robotics applications. We have also proposed our learning-based loop closure detection network \textit{Loop-Net}. The network is heavily relied on our previous work FG-Net \cite{liu2020fg, liu2021fg, liu2022fg}. We use the classification loss to judge if there exists overlaps between the two LiDAR scans. And we use the regression loss to predict the relative angle between the two LiDAR scans.

\section{The Integrated System for Local Building Inspections at Hong Kong}
    Finally, we have successfully utilized the above LiDAR SLAM system to conduct building inspections and obtain the surveillance locally at Hong Kong. The System is illustrated in Fig. \ref{fig_System_Inspection}. It has been demonstrated by experiments that our proposed LiDAR-Inertial simultaneous localization and mapping system has great capacity to generate high-quality map, which is of great help to conduct autonomous motion planning to generate feasible path. The constructed 3D map by the LiDAR SLAM system is also of great significance to the further applications such as 3D building reconstructions, 3D model analysis, and also building targets or defects localization and detection.
\section{Conclusions}
In this work, we have proposed an integrated improved LiDAR-Inertial simultaneous localization and mapping system for unmanned ground vehicle localizations~\cite{liu2022industrial}. We have proposed a systematical design of particle filter-based odometry. By conducting local optimization, we improve the robustness and efficiency of previous LiDAR-Inertial SLAM. And by the proposed learning-based loop closure detection method, we can improve the accuracy of the final global map. It can be demonstrated that our proposed LiDAR-Inertial SLAM system shows great performance and robustness in both indoor and outdoor circumstances with mechanical LiDAR or solid-state LiDAR with good accuracy. Tn the future, we want to explore robotic scene parsing with semantic SLAM \cite{liu2022ws3d}.

\addtolength{\textheight}{0cm}   





\bibliographystyle{IEEEtran}
\bibliography{references}

\end{document}